# VacancySBERT: the approach for representation of titles and skills for semantic similarity search in the recruitment domain

**Maiia Y. Bocharova[1)]**
ORCID: https://orcid.org/0009-0004-3875-5019; bocharova.maiia@gmail.com
**Eugene V. Malakhov[1)]**
ORCID: https://orcid.org/0000-0002-9314-6062; eugene.malakhov@onu.edu.ua. Scopus Author ID: 56905389000
**Vitaliy I. Mezhuyev[2)]**
ORCID: https://orcid.org/0000-0002-9335-6131; Scopus Author ID: 24468383200
[1)] Odessa I. I. Mechnikov National University, 2, Dvoryanska Str. Odessa, 65082, Ukraine
[2)] FH JOANNEUM University of Applied Sciences, Werk-VI-Straße 46. Kapfenberg, 8605, Austria

**ABSTRACT**

The paper focuses on deep learning semantic search algorithms applied in the HR domain. The aim of the article is developing a novel approach to training a Siamese network to link the skills mentioned in the job ad with the title. It has been shown that the title normalization process can be based either on classification or similarity comparison approaches. While classification algorithms strive to classify a sample into predefined set of categories, similarity search algorithms take a more flexible approach, since they are designed to find samples that are similar to a given query sample, without requiring pre-defined classes and labels. In this article semantic similarity search to find candidates for title normalization has been used. A pre-trained language model has been adapted while teaching it to match titles and skills based on co-occurrence information. For the purpose of this research fifty billion title-descriptions pairs had been collected for training the model and thirty three thousand title-description-normalized title triplets, where normalized job title was picked up manually by job ad creator for testing purposes. As baselines FastText, BERT, SentenceBert and JobBert have been used. As a metric of the accuracy of the designed algorithm is Recall in top one, five and ten model's suggestions. It has been shown that the novel training objective lets it achieve significant improvement in comparison to other generic and specific text encoders. Two settings with treating titles as standalone strings, and with included skills as additional features during inference have been used and the results have been compared in this article. Improvements by 10 % and 21.5 % have been achieved using VacancySBERT and VacancySBERT (with skills) respectively. The benchmark has been developed as open-source to foster further research in the area.

## 1. INTRODUCTION

Recruitment is a field that has traditionally relied on keyword matching to identify candidates with the relevant skills and experience for a given job opening. As a logical improvement for keyword-based parsers "grammar-based" approaches were introduced.

A grammar-based parser is a type of natural language processing tool that uses grammatical rules to analyze the context of words in a vacancy/resume. This approach tends to provide a more detailed analysis than a keyword-based parser, which only looks for specific words or phrases. By using computational semantics, a grammar-based parser can understand the different meanings of a word or phrase depending on the context in which it is used. However, the drawback of such an approach is that it requires a lot of manual work done by domain experts and does not scale well.

Recent advances in natural language processing have made it possible to automatically learn the representation of words, phrases and sentences in a way that captures their meaning and relationship to one another. This allows for more accurate and efficient semantic similarity comparisons while eliminating the need for expert-created rules.

Titles and requirements specified in the Job Ads provide crucial textual information for Candidate-Vacancy matching.

Although job titles generally follow a certain structure [1], they are often freely structured and could contain some degree of bias.

Standard classification lists with tasks, skills and occupations were established, among which American Occupational Information Network (O*NET) [2] and European Skills, Competences, Qualifications and Occupations (ESCO) [3].

There was some work carried out aiming at normalizing job titles and relating their free form to the standardized job title form [1, 4], [5].

However [4] points out that only 65 % of job titles could be linked to their corresponding normalized taxonomy without ambiguity. This leaves 35 % of Job Ads for which it is of utmost importance to analyze requirements specified in the job description section to map the job to the taxonomy.

To bypass the need for a large hand-labeled dataset, we, following authors of [4, 5] make use of job ads and work experience sections extracted from anonymized resumes to learn the effective representation of titles and skills mentioned in requirements sections of vacancies. To exploit this signal distant supervision settings (in which there is no need for explicitly human-annotated data) are applied in which our semantic similarity model learns to understand the meaning of requirements present in the Job Ad and map them to corresponding job titles. For the test setting collected from governmental job board [6] is used, which in addition to the usual job title and job description section provides a normalized form of the title compliant with the ESCO classification, hand-picked by the Job Ad creator.

The main contributions of this paper are:

1. A novel approach for training a deep neural network to disambiguate the title and determine the normalized title form based on free-form title and skills extracted from job description.

2. Development and making open-source a new dataset consisting of 30k+ data points with title, skills and normalized title counterpart to foster further research in this area.

## 2. LITERATURE ANALYSIS

Learning meaningful vector representation of words, phrases and sentences could be used in many applications [7, 8], [9, 10], [11, 12], [13] – similarity search, recommendations, question answering, classification etc. Because of this, mapping sentences and phrases into a fixed-size, dense vector space continues to draw the extensive attention of researchers

### 2.1. Word embeddings

One approach to learning the representation of words is through the use of word embedding algorithms, such as word2vec [14], GloVe [15] or fasttext [16]. These algorithms take a large corpus of text and learn vector representations for each word, such that words often seen in the same context are mapped to nearby points in the vector space. However, this approach has its limitations and struggles with some aspects of language, such as words with multiple meanings and making representations of phrases, because phrase vectors are computed by just the sum of the individual word vectors.

### 2.2. Pre-trained language models

Lately, large language models like BERT [17], which are capable of learning the complexities and nuances of human language due to context-aware word representations, are learnt through unsupervised language modeling on the large corpora of text (often consisting of billions of words) revolutionized the field. However, the disadvantage of BERT is that it is not suitable to compute independent sentence embeddings [18].

There have been studies which aim to adapt BERT embeddings to be representative at sentence-level. As such – Universal Sentence Encoder [19] and SentenceBERT [18] models were introduced, and trained with supervision leveraging large human-labeled datasets [20, 21]. Those models are dual encoder models consisting of paired encoding models which are trained to map texts in the vector space so that source and target sentences are mapped to the nearby points in the vector space. As to the training objective: the most popular are three tasks - namely regression (cosine similarity between a pair of sentences is computed), triplet (anchor, positive and negative sentence are fed to the network and model is tuned to minimize the distance between positive pair and maximize between the negative one) or ranking objective function with inbatch negative sampling where the model is trained to distinguish between a real positive pair and all other sentence pair combinations.

Therefore, these models are capable of mapping texts into fixed-sized dense vectors while capturing the meaning of the texts.

### 2.3. Embedding techniques and pre-trained language models in the HR domain

Recently there has been an increase in interest towards applying special encoders in the HR-related domain.

In [1] use a large hand-labeled dataset of similar job titles and train a Siamese Recurrent Neural network to project job titles representation into the vector space.

In [22] suggested treating the task as a multi-label text classification problem and using F1 score to evaluate the performance of the classifier.

In [4] train a Siamese network to extract and aggregate features for each token in the job title, using the skills associated with each title as supervision and train a model to predict the presence or absence of the skill in a skip-gram setting.

In [5] use the aggregated representation of skills, produced by the Distributed Bag of Words version of Doc2Vec and train encoder model to map title representation to the Doc2Vec vector of the skills belonging to it.

### 2.4. The formulation of investigation's aim

Using only title embeddings during inference does not account for the contents of the job ad, which could provide valuable elaborations and disambiguate not clearly constructed job titles. Approaching the task as a multilabel text classification problem poses the disadvantage of having a fixed number of classes and suffers from the screwed distribution problem.

Thus, the aim of this research is developing a novel approach to training a Siamese network to link the skills mentioned in the job ad with the title. For evaluating which it is necessary to investigate two settings: the first is to treat titles as standalone strings, the second one includes the skills as additional features during inference.

### 3. DESIGN AND METHODS OF EXPERIMENT

To minimize the computations costs and time needed for training, model weights should not be not initialized randomly, but are initialized from pretrained BERT-base weights. As such the encoder layer of VacancySBERT is initialized from pretrained BERT-base weights. The base version over the large one was chosen for efficiency reasons. The original authors of BERT used [SEP] token to separate the first and second segments which they later used for a Next Sentence Prediction training objective. In spite of the fact that the pretrained model has not been trained on more than two segments, it is demonstrated by the conducted experiments described in the later section that VacancySBERT is capable of generalizing to multiple segments per input sequence.

A new [SKILL] token is introduced, which is initialized from one of the unused BERT tokens.

This [SKILL] token should capture the meaning of the skill which comes after it. Representations of all [SKILL] tokens are being averaged and fed to the linear pooling layer to reduce dimensions.

The overview of the model's architecture is visualized in Fig. 1.

The model is trained to minimize the distance between the title embedding and corresponding aggregated skills representation, using Multiple Negative Ranking loss with in-batch negatives [24].

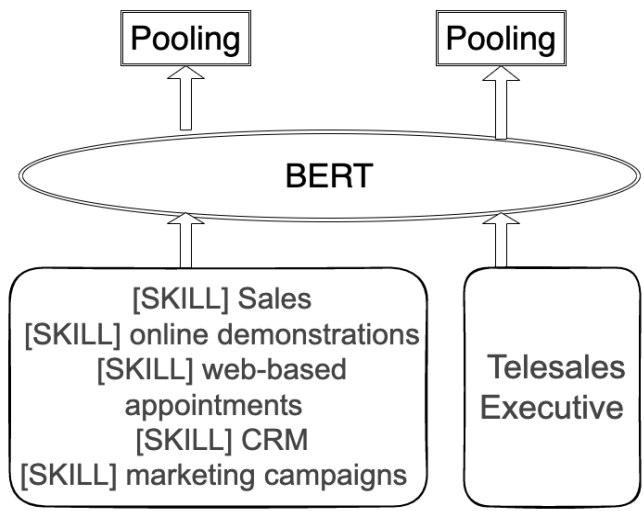

*Fig. 1.* **Model architecture**
*Source*: compiled by the authors

### 4. EXPERIMENTS

#### 4.1. Benchmark dataset

Following [4] we collect a dataset of Job Ads from the big governmental job board website myfuturejobs.gov.ua [6], which not only contains the job titles and corresponding descriptions, but also a normalized ESCO compliant job title form, selected by the creator of each post and use it for evaluation purposes. We carefully remove job ads containing non-English descriptions, the statistics of the final dataset used for testing is shown in Fig. 2. This dataset is made open-source [23].

As we can see from Fig. 2 the collected dataset is imbalanced, with jobs mainly posted for the "Professional" and "Technical and associate professionals" occupations family, while "Skilled agricultural, forestry and fishery workers" as well as "Armed forces occupations" are almost not represented at all.

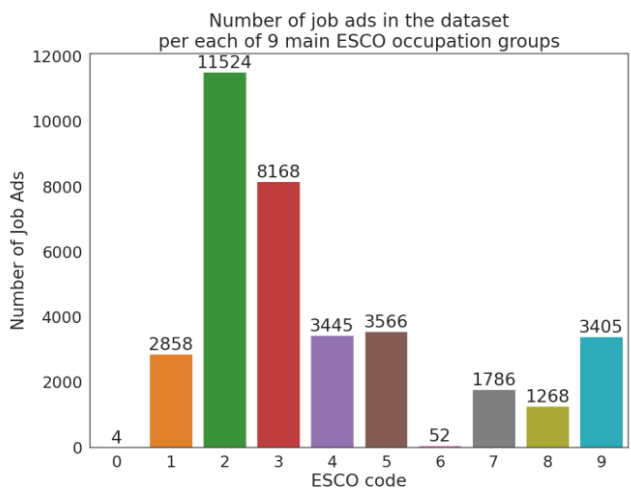

*Fig. 2*. **Number of job ads in the collected benchmark dataset**
*Source*: compiled by the authors

* where ESCO code corresponds to:
0 – Armed forces occupations;
1 – Managers;
2 – Professionals;
3 – Technical and associate professionals;
4 – Clerical support workers;
5 – Service and sales works;
6 – Skilled agricultural, forestry and fishery workers;
7 – Craft and related trades workers;
8 – Plant and machinery operators and assemblers;
9 – Elementary occupations

When working with imbalanced datasets, classification tasks can be challenging as the minority class can be easily overshadowed by the majority class. Ranking objectives, on the other hand, can be a better option for such datasets. Ranking is a technique that focuses on the relative ordering of samples rather than absolute classification. It aims to distinguish between a real positive pair and all other pair combinations. This approach is effective for imbalanced datasets because it does not require an equal number of positive and negative examples, making it more flexible and applicable to real-world scenarios where the minority class is prevalent.

Additionally, ranking objectives can also prioritize the correct ranking of misclassified examples, thereby improving the overall model performance.

**4.2. Training data collection and preprocessing**

We select a large training corpus of ~10 million English vacancies, collected from different US and UK job boards and ~40 million anonymized work history sections from Daxtra's internal Data Lake. While training a multilingual models is possible we focus on English language (language is determined by pretrained fasttext language detector [16])

The statistics of the collected dataset is shown in Table 1.

*Table 1*. **Large-scale training dataset**

| Source | Number of unique descriptions | Number of unique titles | Average number of descriptions per title |
|---|---|---|---|
| Resume | 40M | 6.5M | 6.17 |
| Vacancies | 10M | 4.8M | 2.08 |

*Source*: compiled by the authors

Both resumes and vacancies contain very diverse job titles. As could be noticed, resumes contain much cleaner job titles compared to the ones present in vacancies, since in advertisements there is usually a lot of extra information like job perks, locations etc.

Using regular expressions we remove URLs, emails and phone numbers. Afterwards to clean the collected vacancies of irrelevant information we apply an in-house model for context-aware binary classification of the sentences in terms of them either belonging to job description and qualification requirements sections or extra company and benefits relating parts to ensure that we extract the skills only from relevant sections (and prevent possibility of extracting "skills" from the sections which contain descriptions of benefits advertised and other irrelevant for our task information).

Since the same word, for example "Python", could be written in several variations (Python, python, PYTHON etc) we lowercase our corpus to make the vocabulary more representative.

A proprietary algorithm, operating on span levels is utilized for skill identification to extract the skills to use for self-supervision.

**4.3. Training details**

Analysis of the extracted skills shows that most job ads contain from 10 to 100 skills, with only 8.6 % containing less than 10 and only ~12.4 % containing more than 100 skills.

Number of skills per job ad in shares is shown in Fig. 3.

The maximum length of the concatenated skills is set to 128 tokens as we find that in most job ads this length is enough to successfully include all the skills present there.

Multiple Negative Ranking loss [24] with in-batch negatives is used and a batch size of 32 is set. Model is trained for one epoch, while monitoring the performance on a validation subset is conducted.

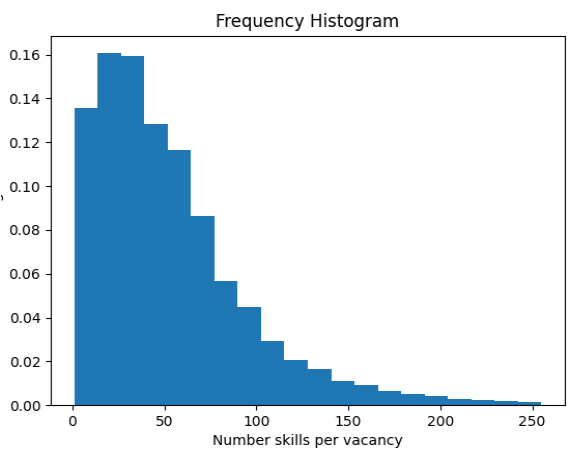

*Fig. 3.* **Number of skills per job ad**
*Source*: compiled by the authors

The Recall@N as a metric is employed. This metric measures if a correct item was retrieved among top N model's recommendations. A high Recall@N means that the system is able to accurately identify the relevant item among the candidates.

The results of the experiment can be seen in Table 2.

It can be noted that adding skills even to the pretrained SBERT improves predictive power of the model.

To compare the proposed model with the best competitor (JobBERT) and to find the improvement (Δ) for the two settings (with or without using skills as additional features) the following formula is used:

$$\Delta = \left( \sum_{N=1,5,10} \frac{R@N_{V.B.} - R@N_{J.B.}}{R@N_{J.B.}} \times 100\% \right) / 3 \quad (1)$$

where R@N corresponds to Recall@1, Recall@5, Recall@10 in Table 2, V.B. – corresponds to VacancySBERT (with or without skills) in Table 2 and J.B. to JobBERT respectively.

As a result of calculations, improvements by 10% and 21,5 % are achieved using VacancySBERT and VacancySBERT (with skills) for comparison respectively.

Thus, a new training approach allows us to improve results significantly. Using skills as additional features during inference increases the effect obtained.

*Table 2.* **Titles normalization**

| Model | Recall@1 | Recall@5 | Recall@10 |
|---|---|---|---|
| FastText | 0.134 | 0.223 | 0.279 |
| BERT* | 0.142 | 0.180 | 0.254 |
| SBERT | 0.219 | 0.320 | 0.444 |
| SBERT (with skills) | 0.233 | 0.361 | 0.50 |
| JobBERT** | 0.225 | 0.386 | 0.46 |
| VacancySBERT (without skills) | 0.271 | 0.402 | 0.489 |
| VacancySBERT (with skills) | 0.301 | 0.425 | 0.556 |

*Source*: compiled by the authors

\* We calculate the embedding of the title by averaging the embeddings of all the words

\*\* Metrics are based on the numbers provided in [3], since the model was not open-source

## 5. CONCLUSIONS

This paper presented a novel approach for job title normalization that builds on the premise that skills are the essential components defining a job. It has been demonstrated that proposed approach - VacancySBERT – which uses distant supervision to learn to represent job vacancy titles based on skills included in their description.

Thus, the proposed approach allows for more accurate job title normalization, as it can identify similar job titles based on the skills required, even if the job titles themselves are different. This approach avoids the need for manual data labeling, which can be costly and time-consuming, especially given the rapidly changing trends in the job market.

The advantages of the proposed approach can be seen from significant improvements in semantic text similarity search. Indeed, the new approach outperforms state-of-the-art encoders for semantic text similarity for the task of job titles normalization by 10-21.5 % depending on the usage of skills during inference.

Thus, demonstrating the effectiveness of our approach on this challenging task, it has been shown that VacancySBERT can be used to build more accurate and scalable job title normalization systems that can keep up with the constantly evolving job market.

## 6. FUTURE WORK

Possible way to improve the obtained results is to develop a better negative sampling strategy as well as score the skills in terms of their importance and representativeness. TF-IDF could be used for that.

Also, as a future work we plan to extend the approach for multilingual settings.

## REFERENCES

1. Neculoiu, P., Versteegh, M. & Rotaru, M. "Learning text similarity with siamese recurrent networks". *Proceedings of the 1st Workshop on Representation Learning for NLP*. Berlin: Germany. 2016. p. 148–157. DOI: https://doi.org/10.18653/v1/W16-1617.

2. "O*NET OnLine". – Available from: www.onetonline.org. – [Accessed: 8, January, 2022].

3. "ESCO: European skills, competences, qualifications and occupations". *EC Directorate E.* 2017. – Available from:: https://esco.ec.europa.eu/en. – [Accessed 8 January, 2022].

4. Decorte, J.-J., Van Hautte, J., Demeester, T. & Develder, C. "JobBERT: Understanding job titles through skills". 2021. DOI: https://doi.org/10.48550/arXiv.2109.09605.

5. Zbib, R., Lacasa Alvarez, L., Retyk, F. et al. "Learning job titles similarity from noisy skill labels". 2022. DOI: https://doi.org/10.48550/arXiv.2207.00494.

6. "The malaysian national employment portal for all job seekers and employers". – Available from: https://candidates.myfuturejobs.gov.my/search-jobs. – [Accessed: 8, January, 2022].

7. Malakhov, E., Shchelkonogov, D. & Mezhuyev, V. "Algorithms for classification of mass problems of production subject domains". *ACM Proceedings of the 2019 8th International Conference on Software and Computer Applications.* Penang: Malaysia. 2019. p. 149–153. DOI: https://doi.org/10.1145/3316615.3316676.

8. Mezhuyev, V., Malakhov, E. & Shchelkonogov, D. "The method and algorithms to find essential attributes and objects of subject domains". *International Conference on Computer, Communication and Control Technology (I4CT).* Kuching, Sarawak: Malaysia. 2015. p. 310–314. DOI: https://doi.org/10.1109/I4CT.2015.7219587.

9. Al-Emran, M., Abbasi, G. A. & Mezhuyev, V. "Evaluating the impact of knowledge management factors on M-Learning adoption: A deep learning-based hybrid SEM-ANN approach". *Recent Advances in Technology Acceptance Models and Theories. Studies in Systems. Decision and Control. Springer*. 2021; 335: 159–172. DOI: https://doi.org/10.1007/978-3-030-64987-6_10.

10. Al-Emran, M., Mezhuyev, V. & Kamaludin, A. "Students' perceptions towards the integration of KM processes in M-learning systems: A preliminary study". *International Journal of Engineering Education*. 2018; Vol. 34. No. 2(A): 371–380.

11. Al-Emran, M., Mezhuyev, V. & Kamaludin, A. "An innovative approach of applying knowledge management in M-learning application development: A pilot study". *International Journal of information and communication technology education*. 2019; 15(4): 94–112. DOI: https://doi.org/10.4018/IJICTE.2019100107.

12. Mezhuyev, V., Sadat, S. M. N., Rahman, M. A., Refat, N. & Asyhari A. T. "Evaluation of the likelihood of friend request acceptance in online social networks", *IEEE Access*. 2019; 7: 75318–75329. DOI: https://doi.org/10.1109/ACCESS.2019.2921219.

13. Rahman, M. A., Mezhuyev, V., Bhuiyan, M. Z. A., Sadat, S. M. N., Zakaria, S. A. B. & Refat, N. "Reliable decision-making on accepting friend requests in online social networks". *IEEE Access. Cyber-Physical-Social Computing and Networking*. 2018; 6 (1): 9484–9491. DOI: https://doi.org/10.1109/ACCESS.2018.2807783.

14. Mikolov, T., Chen, K., Corrado, G. & Dean, J. "Efficient estimation of word representations in vector space". 2013. DOI: https://doi.org/10.48550/arXiv.1301.3781.

15. Pennington, J., Socher, R. & Manning, C. D. "Glove: Global vectors for word representation". *Proceedings of the 2014 conference on empirical methods in natural language processing (EMNLP)*. Doha: Qatar. 2014. p. 1532–1543. DOI: https://doi.org/10.3115/v1/D14-1162.

16. Bojanowski, P., Grave, E., Joulin, A. & Mikolov, T. "Enriching word vectors with subword information". *Transactions of the Association for Computational Linguistics*. 2016; 5. DOI: https://doi.org/10.48550/arXiv.1607.04606.

17. Devlin, J., Chang, M.-W., Lee, K. & Toutanova, K. "BERT: Pre-training of deep bidirectional transformers for language understanding". 2018. DOI: https://doi.org/10.48550/arXiv.1810.04805.

18. Reimers, N. & Gurevych, I. "SentenceBERT: Sentence embeddings using siamese BERT networks". *Processing and the 9th International Joint Conference on Natural Language Processing (EMNLP-IJCNLP)*. Hong Kong: China. 2019. p. 3982–3992. DOI: https://doi.org/10.48550/arXiv.1908.10084.

19. Cer, D., Yang, Y., Kong, S., Hua, N. et al. "Universal Sentence Encoder". 2018. DOI: https://doi.org/10.48550/arXiv.1803.11175.

20. Bowman, S. R., Angeli, G., Potts, C. & Manning, C. D. "A large annotated corpus for learning natura l language inference". *In Proceedings of the 2015 Conference on Empirical Methods in Natural Language Processing (EMNLP)*. 2015. DOI: https://doi.org/10.48550/arXiv.1508.05326.

21. Williams, A., Nangia, N. & Bowman, S. "A broad-coverage challenge corpus for sentence understanding through inference". *In Proceedings of the 2018 Conference of the North American Chapter of the Association for Computational Linguistics: Human Language Technologies*. 2018; 1: 1112–1122. DOI: https://doi.org/10.48550/arXiv.1704.05426.

22. Tran, H. T., Vo, H. H. P. & Luu, S.T. "Predicting job titles from job descriptions with multi-label text classification". *8th NAFOSTED Conference on Information and Computer Science (NICS)*. 2021. DOI: https://doi.org/10.48550/arXiv.2112.11052.

23. Dataset used in the paper "VacancySBERT…". – Available from: https://github.com/maiiabocharova/VacancySBERT. – [Accessed: 8, January, 2022].

24. Henderson, M., Al-Rfou, R., Strope, B et al. "Efficient natural language response suggestion for smart reply". 2017. DOI: https://doi.org/10.48550/arXiv.1705.00652.

# VacancySBERT - підхід до представлення назв посад та навичок для семантичного пошуку в домені підбору персоналу

**Бочарова Майя Юріївна[1)]**
ORCID: https://orcid.org/0009-0004-3875-5019; bocharova.maiia@gmail.com
**Малахов Євгеній Валерійович[1)]**
ORCID: https://orcid.org/0000-0002-9314-6062; eugene.malakhov@onu.edu.ua. Scopus Author ID: 56905389000
**Межуєв Віталій Іванович[2)]**
ORCID: https://orcid.org/0000-0002-9335-6131; Scopus Author ID: 24468383200
[1)] Одеський національний університет імені І. І. Мечникова, вул. Дворянська, 2. Одеса, 65082, Україна
[2)] Университет прикладних наук FH JOANNEUM. Werk-VI-Straße 46.  Капфенберг, 8605, Австрія

## АНОТАЦІЯ

Стаття присвячена алгоритмам семантичного пошуку з глибоким навчанням, що застосовуються у сфері управління персоналом. Метою дослідження є вдосконалення та розширення різноманітних підходів до нормалізації назв, написаних у вільній формі, для зіставлення із заздалегідь визначеною стандартною таксономією. Завдання дослідження - запропонувати нову навчальну задачу для великої мовної моделі та навчити її відображати назви посад у вільній формі та навички, які пов'язані із назвою посади, у векторний простір таким чином, щоб назви посад, які мають спільне значення, знаходилися близько один до одного. Процес нормалізації назв посад може ґрунтуватися або на класифікації, або на порівнянні схожості. У той час як алгоритми класифікації намагаються віднести вибірку до заздалегідь визначеного набору категорій, алгоритми пошуку подібності застосовують більш гнучкий підхід, оскільки вони призначені для пошуку зразків, схожих на задану вибірку запиту, не вимагаючи заздалегідь визначених класів і міток. Враховуючи це, для пошуку кандидатів на нормалізацію назв посад ми будемо використовувати пошук за семантичною схожістю. Попередньо навчена мовна модель адаптується під час навчання для зіставлення назв посад і навичок на основі інформації про спільні входження. Для цього дослідження було зібрано близько 50 мільйонів пар "назва посади-опис" для навчання моделі та 33 тисячі триплетів "назва посади-опис-нормалізована назва посади" для тестування, де нормалізована назва посади була підібрана вручну укладачем оголошення про роботу. В якості базових моделей використано FastText, BERT, SentenceBert та JobBert. Як метрику точності розробленого алгоритму використано показник Recall у 3, 5 та 10 найкращих пропозиціях моделі. Показано, що нова мета навчання дозволяє досягти значного покращення порівняно з іншими загальними та специфічними кодувальниками тексту. Результати проаналізовано та використано для формулювання висновків та пропозицій щодо подальшої роботи. Датасет, який використовувався для тестування моделей оприлюднено задля сприяння подальшим дослідженням у цій галузі.

**Ключові слова:** обробка природної мови; векторне представлення документів; семантичний пошук; машинне навчання; векторне представлення речень; глибокі нейронні мережі; інтелектуальна обробка даних

## ABOUT THE AUTHORS

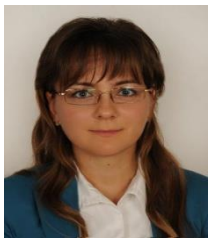

**Maiia Y. Bocharova** - Postgraduate, Department of Mathematical Support of Computer Systems, Odessa I.I. Mechnikov National University, 2, Dvoryanska Str. Odessa, 65082, Ukraine
ORCID: https://orcid.org/0009-0004-3875-5019; bocharova.maiia@gmail.com
*Research field*: Natural Language Processing; similarity search; machine learning; data mining, software engineering

Бочарова Майя Юріївна - аспірант, кафедра Математичного забезпечення комп'ютерних систем. Одеський національний університет імені І. І. Мечникова, вул. Дворянська, 2. Одеса, 65082, Україна

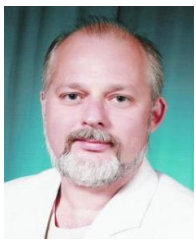

**Eugene V. Malakhov** - Doctor of Engineering Sciences, Professor, Head of Department of Mathematical Support of Computer Systems, Odessa I. I. Mechnikov National University, 2, Dvoryanska Str. Odessa, 65082, Ukraine
ORCID: https://orcid.org/0000-0002-9314-6062; eugene.malakhov@onu.edu.ua. Scopus Author ID: 56905389000
*Research field*: Databases theory; metamodeling; the methods of data mining and other data structuring; data processing methods

Малахов Євгеній Валерійович - доктор технічних наук, професор, завідувач кафедри Математичного забезпечення комп'ютерних систем. Одеський національний університет імені І. І. Мечникова, вул. Дворянська, 2. Одеса, 65082, Україна

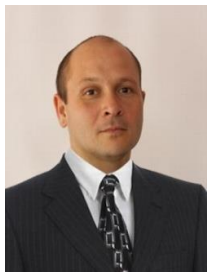

**Vitaliy I. Mezhuyev** - Doctor of Engineering Sciences, Professor, FH Joanneum
Werk-VI-Straße 46. Kapfenberg, 8605, Austria
ORCID: https://orcid.org/0000-0002-9335-6131; vitaliy.mezhuyev@fh-joanneum.at. Scopus ID: 24468383200
*Research field*: Formal methods; metamodeling; safety modeling and verification of hybrid software systems, and the design of cyber-physical systems

Межуєв Віталій Іванович - доктор технічних наук, професор Інституту промислового менеджменту. Університет прикладних наук FH JOANNEUM. Капфенберг, Австрія